\newif\ifconfpaper
\confpapertrue 

\newif\ifconfletter

\ifconfpaper
\documentclass[letterpaper, 10 pt, conference]{ieeeconf}
\IEEEoverridecommandlockouts
\fi

\ifconfletter
\documentclass[letterpaper, 10 pt, journal, twoside]{IEEEtran}  
\fi

\usepackage{graphicx}
\usepackage{amsmath}
\usepackage{amsfonts}
\usepackage{bm}
\usepackage{booktabs}

\usepackage{color}
\usepackage{url}
\usepackage[dvipsnames]{xcolor}

\newcommand{\secref}{Section~\ref}
\newcommand{\figref}{Figure~\ref}

\newcommand{\etal}{\textit{et al}.}
\newcommand{\ie}{\textit{i}.\textit{e}.}
\newcommand{\eg}{\textit{e}.\textit{g}.}

\ifconfletter
\title{}
\else
\title{\LARGE \bf Task-grasping from human demonstration}
\fi
\ifconfletter
\author{%
\thanks{Manuscript received: Month, dd, year; Revised Month, dd, year; Accepted Month, dd, year.}
\thanks{This paper was recommended for publication by Editor XXX upon evaluation of the Associate Editor and Reviewers' comments.
}
\thanks{$^{1}$All authors are with Applied Robotics Research, Microsoft, Redmond, WA, USA
        {\tt\footnotesize Kazuhiro.Sasabuchi@microsoft.com}}%
\thanks{Digital Object Identifier (DOI): see top of this page.}
}
\else
\author{Daichi Saito$^{1,2}$, Kazuhiro Sasabuchi$^{1}$, Naoki Wake$^{1}$, Jun Takamatsu$^{1}$, Hideki Koike$^{2}$, and Katsushi Ikeuchi$^{1}$%
\thanks{$^{1}$Authors are with Applied Robotics Research, Microsoft, Redmond, WA, USA
        {\tt\small Kazuhiro.Sasabuchi@microsoft.com}}%
\thanks{$^{2}$Authors are with Tokyo Institute of Technology {\tt\small saito.d.ah@m.titech.ac.jp}}
}
\fi
\ifconfletter
\fi

\ifconfpaper
\begin{document}
\fi

\maketitle

\ifconfpaper
\thispagestyle{empty}
\pagestyle{empty}
\fi

\begin{abstract}
A challenge in robot grasping is to achieve task-grasping which is to select a
grasp that is advantageous to the success of tasks before and after grasps. One of the frameworks to address this difficulty is Learning-from-Observation (LfO), which obtains various hints from human demonstrations. This paper solves three issues in the grasping skills 
in the LfO framework:
1) how to functionally mimic human-demonstrated grasps to robots with limited grasp capability, 2) how to coordinate grasp skills with reaching body mimicking, 3) how to robustly perform grasps under object pose and shape uncertainty.
A deep reinforcement learning using contact-web based rewards and domain randomization of approach directions is proposed to achieve such robust mimicked grasping skills.
Experiment results show that the trained grasping skills can be applied in an LfO system and executed on a real robot. In addition, it is shown that the trained skill is robust to errors in the object pose and to the uncertainty of the object shape, and can be combined with various reach-coordination.
\end{abstract}

\ifconfletter
\fi

%
\ifconfletter
\IEEEpeerreviewmaketitle
\fi

\section{Introduction}
\label{introduction}
%
%
%
%
\ifconfletter
\IEEEPARstart{T}{he} first sentence of the introduction.
\else
\fi

\begin{figure}[t]
      \centering
      \includegraphics[width=\columnwidth]{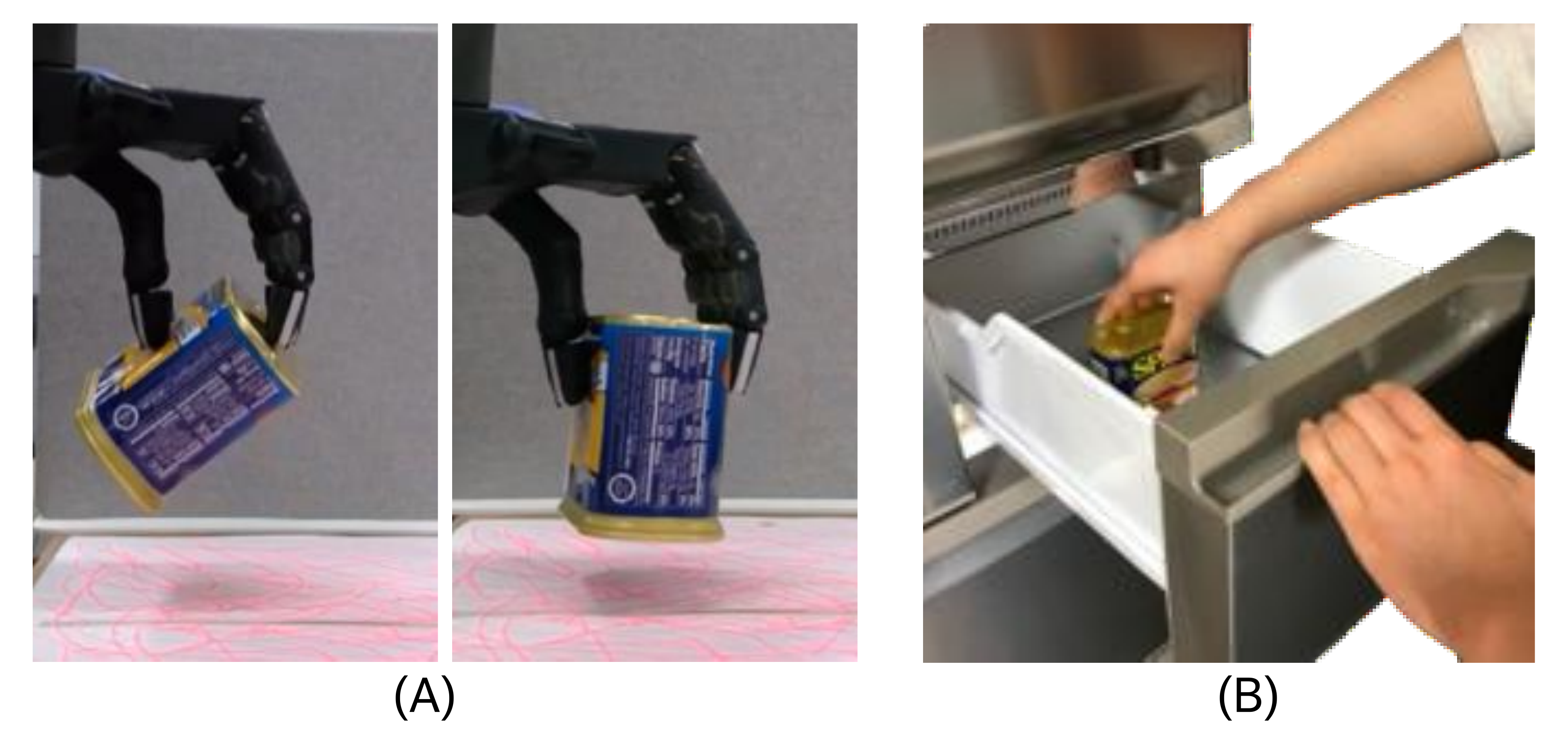}
	 \caption{(A) An example of a grasp acceptable for bin-picking but not for task-grasping (left) versus a grasp acceptable for task-grasping (right). The left is difficult to control the placing task of the object. (B) An example of a grasp constrained by the reaching body. In the dual-arm operation, the left arm is in front of the drawer to perform a pulling task, and the body cannot grasp from the front of the object. Therefore, the grasp is constrained to approach the object from the right.}
      \label{fig:main}
   \end{figure}

The study of robot grasps has evolved in the industrial field.
Yet, a challenge is to achieve context-optimal grasping. The problem of selecting a grasp that is advantageous to the success of tasks before and after grasping is called {\em task-grasping}. The parameters that determine the optimality of a grasp often involve a higher-level understanding of the tasks to be performed. For example, \figref{fig:main}~(A)-left shows a case where a pick acceptable for bin-picking is achieved with a dexterous hand, but does not fulfill the grasp in terms of task-grasping: grasp in a way so that is easy to place the object neatly in a household environment.
The gap between conventional bin-picking and task-grasping lies in the difficulty of understanding the higher-level knowledge about the adjacent tasks.

One of the frameworks to address this difficulty is Learning-from-Observation (LfO), in which a human provides manipulation instruction to a robot through a one-shot demonstration~\cite{ikeuchi1994toward, wake2021learning}. We have previously shown that the LfO framework can be used to teach the humanoid robot implicit high-level knowledge for the task, such as the semantic and physical constraints that exist between the environment and the grasping object, and the reaching body posture that is less likely to conflict with the environment~\cite{sasabuchi2020task}. Planning grasps based on these implicit knowledge has the potential to realize task-grasping.

The LfO-based task-grasping problem can be divided into two sub-problems: 1) developing grasping skills which is capable of leveraging the implicit knowledge that is obtained from human-demonstrated grasps but is also robust against stochasticity such as recognition errors, and 2) the actual obtaining of the implicit knowledge through real time recognition of the one-shot human demonstration. 

For sub-problem 1, recent reinforcement learning techniques 
have been shown to be effective for improving grasping skill robustness,
as simulators can generate various error
patterns which imitate recognition errors, and the robot can be trained under the different error
conditions~\cite{valarezo2021natural, mandikal2021learning, merzic2019leveraging}. However, 
while previous works have achieved robust picks, the implicit knowledge about adjacent tasks obtained from human-demonstrated grasps cannot be leveraged for two reasons: First, previous reward designs do not guarantee-to-mimic human grasps especially when the robot's fingers have limited capability.
Second, previous works have not considered the coordination between the grasp and the reaching body (the approach direction of the grasp toward an object),
due to the assumption that the robot arm can reach from the front of the object. This is not always the case in a complex task sequence
as shown in \figref{fig:main}~(B) and~\cite{sasabuchi2020task}.
Thus, this paper focuses on solving sub-problem 1 by proposing a reward design which guarantees to functionally mimic human grasps using a so-called contact-web expression, and achieve the coordination with the reaching body by training with domain randomization of approach directions.

For sub-problem 2, Kang~\etal~proposed a system to determine the grasp strategy from a human demonstration
~\cite{kang1997toward}. However, due to the lack of solving sub-problem~1, the grasping was vulnerable to errors in the position and orientation of the object at runtime.

The contributions of this paper are as follows:
\begin{itemize}
\item Contact-web based reward: Grasping skills necessary for task-grasping are achieved by using contact-web expressions, which can distinguish grasps by defining the desired contact locations and force directions onto an object. A contact-web based reward in reinforcement learning is proposed to achieve these skills with robustness.
\item Domain randomization of approach directions: Coordination between the grasp and the reaching body is achieved through training-in-simulation by randomizing the "approach directions of the grasp toward an object."
\item Task-grasping from human demonstration: The proposed grasping skills are trained in simulation and then applied in an LfO system and executed on a real robot. Execution results are shown with three representative grasps which functionally mimic human grasps.
\end{itemize}

%

\section{Related Works}
\label{related_works}


In the early years, grasping was more of an analysis research of how to formulate stable grasps~\cite{bicchi2000robotic}. 
After the early years, grasping became more of a kinematic planning problem of how to obtain stable grasps with a specific robot hardware as well as how to reach such stable grasp states from a pre-grasp state~\cite{brost1988automatic}. In these approaches, mostly, the models of the objects to grasp were assumed to be known for calculating the kinematics.
With the development of the deep neural network techniques, several grasp detection methods for unknown objects using convolutional neural network have been proposed~\cite{lenz2015deep, redmon2015real, wang2016robot}. However, these methods were vulnerable to errors in the estimated object pose because of the open-loop control. To solve this problem, several end-to-end systems have been proposed to stably control gripper directly from visual input~\cite{levine2016end, levine2018learning, kalashnikov2018scalable, rao2020rl}. In addition, multi-modal grasping methods combining vision and tactile have been proposed~\cite{calandra2018more, hogan2018tactile}.
These methods are very promising in terms of providing a stable grasp.
However, they ignore the context of tasks and aim for the so-call bin-picking instead of task-grasping. Thus, they cannot achieve the task-grasp along with the task situation and environment, which can be obtained and hinted through human demonstration for optimizing a given task sequence~\cite{sasabuchi2020task}.

In parallel with the effort on stable grasping, researchers also explored the methods to determine the optimal grasping strategy for a given task. Cutkosky's pioneering work proposed a taxonomy of grasping and a rule-based determination method for a given task~\cite{cutkosky1989grasp}. 
Kang~\etal~proposed a system to determine grasping strategy from a human demonstration; they reorganize Cutkosky's taxonomy 
and derived a learning-from-observation system to determine the grasp type~\cite{kang1997toward}. However, even if the grasping is obtained, it is vulnerable to errors in the position and orientation of the object at runtime, and stochastic processes must be introduced. 
To this end, most of the recent studies have applied reinforcement learning to train grasping skills robust against recognition errors~\cite{merzic2019leveraging, pmlr-v100-wu20a, koenig2021tactile, rajeswaran2017learning, valarezo2021natural, mandikal2021learning}.
Especially, \cite{rajeswaran2017learning, valarezo2021natural, mandikal2021learning} can train grasping skills which mimic the human grasp shape.
Yet, it is impractical to mimic human grasp shapes when the robot's fingers have limited capability. In addition, the prior taxonomies are redundant for robots with limited capability.


\section{Method}
\label{method}

This section will first explain the representative grasps required in the LfO-based task-grasping problem, and how they can be expressed using the contact-web expression. Then, using these expressions, we explain how we can train and develop robust grasping skills based on each grasp.

\subsection{defining grasps and contact-webs}

\begin{figure}[t]
      \centering
      \includegraphics[clip,width=\columnwidth]{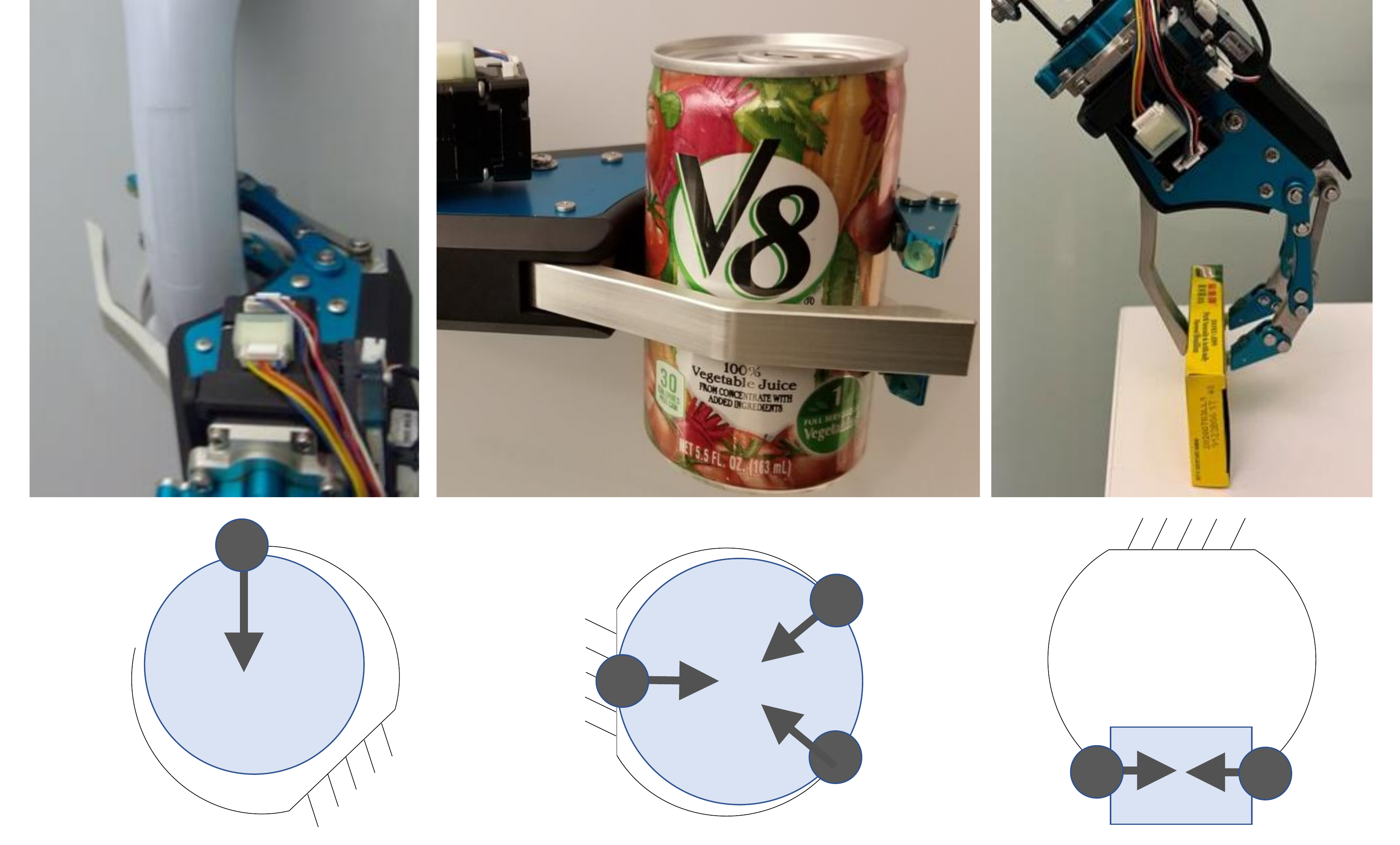}
	 \caption{Three representative grasps performed with a one-degree-of-freedom gripper (top) and the corresponding contact-webs (bottom). Black circles and arrows indicate the contact points and the desired force directions, respectively. The black arcs illustrate the gripper.}
      \label{fig:threeg}
  \vspace{-3mm}
   \end{figure}

Reproducing human grasps is essential in task-grasping, and especially when human knowledge must be integrated to overcome the high-level decision making. 
Since the human can only grasp within the possible shape producible by his or her fingers, the number of grasps must be finite, therefore, we can define a skill set beforehand, and the high-level decision making can decide which skill to use.

Yet, the difficulty in defining the different grasps, is that, the definition of the grasp must make sense to the robot. Defining grasps by the shape of the human grasp will not make sense to robotic fingers which capabilities are usually limited compared to the human fingers; in the worst case with only two-opposing fingers. Due to the limited capability of the fingers, it is impractical or even impossible to generate a finger configuration that mimics exactly like a human grasp shape. 
Instead of looking at the grasps by shape, we propose looking at grasps by functions using the closure theory perspective.
The theory generalizes to a more variety of finger configurations, and can mimic human grasps in terms of the function of the grasp, \ie, the requirements between the fingers and object/tools to achieve tasks after grasping.

When we categorize grasps by closure, we are able to define three 
representative grasps~\cite{yoshikawa1999passive}. That is, passive-form closure, passive-force closure, and active-force closure. The difference between form and force closure is whether some extra force is required to be produced by the fingers in order to resist external force such as gravity. Taking this definition broadly, we can also say a loose-hooking against a handle (\figref{fig:threeg}(A)) is also a form closure in that, the handle cannot get away, and gravity is self compensated by the environment and does not require extra force. Some may argue this is not a grasp, however, in terms of initiating connection with a tool, this is common and in-fact the second most type of connections done by a human \cite{VERGARA2014225}, and thus cannot be ignored.
The difference between passive-force and active-force closure is whether fingertips can actively produce the force or not. The passive closure allows a more firm grasp, while the active closure can produce precise control against external force.

\figref{fig:threeg} shows that even for a simple one-degree-of-freedom gripper, the grasp can be distinguished in terms of the mimicked functions.
Even just three, distinguishing these is critical for task-grasping.

These grasps by closure can be defined by the relation between the grasping object and the contact locations of the virtual fingers~\cite{arbib1985coordinated} (fingers that coordinate in movement are counted as one virtual finger; for example, index-middle-ring-little are four fingers but can be counted as one virtual finger for the grasps in this paper), as well as the contact force direction produced (which should be perpendicular to the object's surface). The active-force closure can be expressed as appropriately locating $N (\geq 2)$ contacts on the object using the fingertips of the virtual-fingers. 
The passive-force closure requires appropriately locating $N (\geq 3)$ contact points using the virtual fingers and optionally also the base (e.g., palm) to form a firm closure. 
The passive-form closure wraps around the 
grasped
object by the shape of the grasp 
and can add any $N (\geq 1)$ number of contacts between the object and the form-closed grasp shape. 

We call these appropriate virtual finger/base contact locations (with also information on the desired contact force direction on each location) as contact-webs.

\subsection{train and developing skills}

Our goal is to generate robust task-grasping skills  
by functionally mimicking human grasps.
For the robustness, we use reinforcement learning with domain randomization.
Functional mimicking can be guaranteed if the contact-web is fulfilled with the robot's fingers. 
Therefore, we design rewards that can guide the robot's fingers to become close to the desired contact-web.
Note that a unified reward design about contact-webs is used for all grasps, and each grasp only differ in the desired contact-web to become close against.

The training is done in simulation. After the training, the trained grasping skill policy is directly applied to the real robot without any extra training.
This is achieved with the domain randomization by adding random noises to the initial state of each training episode during the training. These random noises are generated to be similar to the uncertainties the real robot encounters (e.g., recognition errors).
In order to apply the trained policy on the real robot, the state space must contain information that can be acquired by both the simulator and the real robot. In contrast, a reward is only used during the training, so it 
uses information obtained in the simulator but not necessarily obtainable directly with the real robot.

Although this is a grasp training problem, in order to coordinate with the reaching, we define the grasping task as a two-phase problem consisting an approach phase (getting the robot finger contacts closer to the desired contact-web contact location from 15~cm away along the approach direction explained in the next section) and a contact-to-grasp phase (getting the finger contacts close to the desired contact-web contact directions).
The details of the training and reward designs are described below.

\subsubsection{state and action space}

\begin{figure}[t]
      \centering
      \includegraphics[width=5cm]{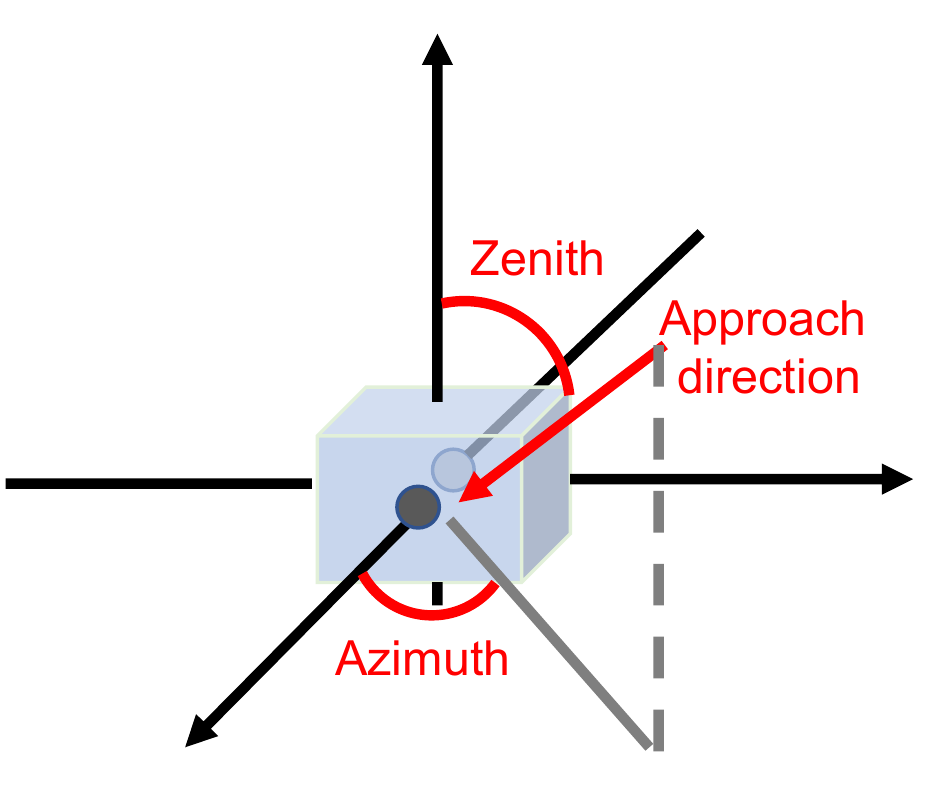}
	 \caption{Illustration of the approach direction defined with zenith and azimuth angles respective to the contact-web.}
      \label{fig:approachdirection}
  \vspace{-3mm}
   \end{figure}

In the state space, we include the actual robot finger positions, the robot wrist orientation, 
and the approach direction of the grasp (represented as zenith, azimuth angle as in \figref{fig:approachdirection}) which is the information used to coordinate with the reach. 
These states are defined respective to the contact-web, which is known in simulation during training, but is recognized when using the trained policy after training with the real robot (see \secref{lfo} for details about after training).
The approach direction is a fixed value during one episode, but is initiated with different values. All other states are updated at every control step (\eg, 100 ms) to the latest state.

In addition, we also add any available sensor feedback with the real robot hardware, such as tactile feedback (e.g., binary values of contacted or not on each finger). If the real hardware does not have any available sensors, states such as the desired finger positions can be used to infer about the contact through desired-actual deviation.
Other information obtainable from simulation (\eg, a vector-formulated contact direction) may be used for the reward calculation, but are not added to the state as is difficult to obtain precisely with a real robot.

For the action space, we include the finger joints of the virtual fingers, as well as changing the wrist pose along the approach direction, \ie, the wrist is allowed to slightly move its position but the wrist orientation is constrained straight around the approach direction.

\subsubsection{fundamentals for the reward design}
\begin{figure}[t]
      \centering
      \includegraphics[width=\columnwidth]{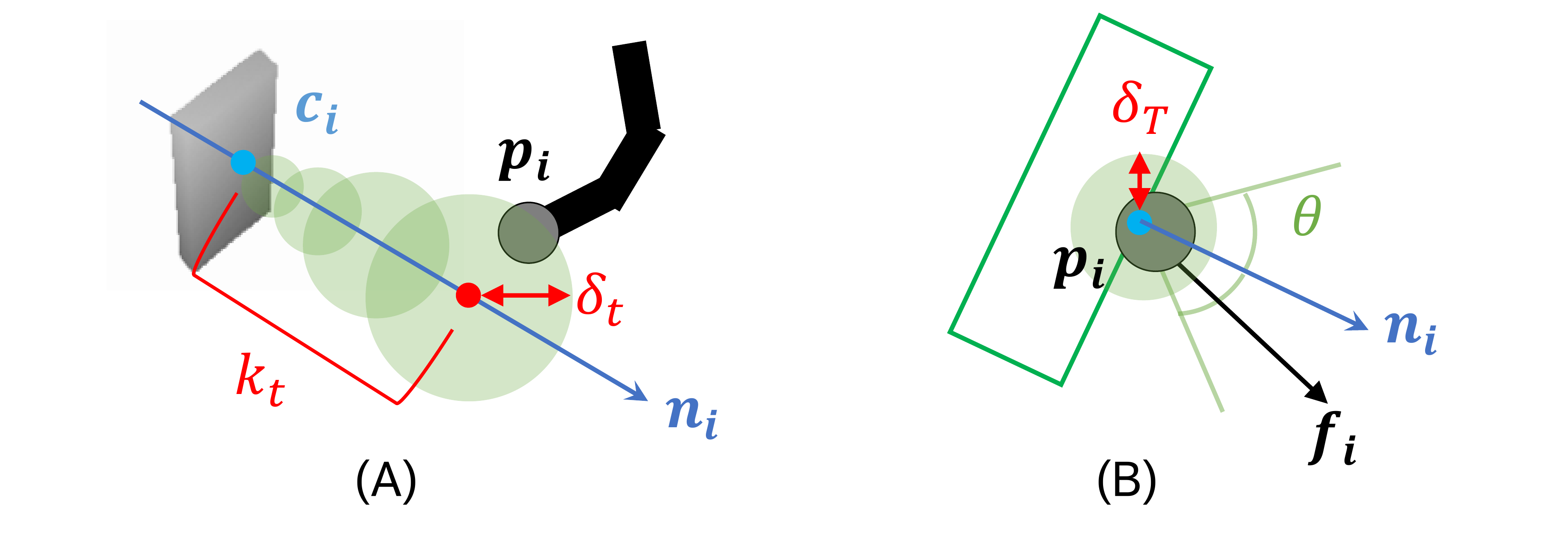}
	 \caption{Illustration of the contact-web based reward design (for simplicity, figure shows only one contact finger). (A) Approach phase. (B) Contact-to-grasp phase.}
      \label{fig:rewards}
  \vspace{-3mm}
   \end{figure}

For each contact~$i$ in the contact-web, a desired contact position $\bm{c}_i$ and a desired contact direction $\bm{n}_i$ is defined.
We minimize the distance $d_p$ between $\bm{c}_i$ and the current position $\bm{p}_i$ of the robot's finger contact, 
but also minimize the cosine distance $d_f$ so that the finger contact direction $\bm{f}_i$ faces $\bm{n}_i$. More precisely, $d_p$ should be minimized by approaching the robot's finger contact along a line which goes through $\bm{c}_i$ and is parallel to $\bm{n}_i$. Thus, for each time step 
$t$, the finger contact should be close enough by $\delta_t$ to some guiding point, which point is at a proportional distance $k_t$ along $\bm{n}_i$ from $\bm{c}_i$ (\figref{fig:rewards}):
\begin{equation}
\label{eq:grasp_task_p}
d_p(\bm{p}_i, t) = \max(\| \bm{p}_i - (\bm{c}_i + k_t\bm{n}_i) \| - \delta_t, 0).
\end{equation}
\begin{equation}
\label{eq:grasp_task_f}
d_f(\bm{f}_i) = \arccos(\bm{f}_i \cdot \bm{n}_i).
\end{equation}
Note that $\bm{c}_i$ and $\bm{f}_i$ is subject to the object's location and pose. During training, the exact state of the object is known in simulation and therefore the above distances can be used for reward calculation.

\subsubsection{the grasping reward}

For each phase (approach and contact-to-grasp), a reward can be designed using the $d_p$, $d_f$ distance metric on each virtual finger. In addition, we can design two type of rewards: a driving reward, and an avoiding penalty. Driving rewards encourage the agent (in this case, the robot) to either achieve or maintain some problem goal. Avoiding penalties discourage the agent from entering an inappropriate state such as failing to maintain a goal. We terminate an episode if any avoiding penalties are met~\cite{peng2018deepmimic}.

The approach phase can be expressed as a driving reward per finger and is formulated as below using an upper-bound baseline $b_p$:
\begin{equation}
\label{eq:approach}
\log(b_p - d_p(\bm{p}_i, t)).
\end{equation}
The approach phase switches to a contact-to-grasp phase once a contact is detected in simulation. 
The contact-to-grasp phase should achieve the target contact direction, \ie, the finger must adjust its contact direction to face a desired direction, while maintaining the contact position.
After contacting at iteration~$T$, the finger should not deviate by more than $\delta_T$. Thus, the contact reward per virtual finger is formulated as below using an upper-bound baseline $b_f$:
\begin{equation}
\label{eq:contact}
\begin{cases} \log(b_f - d_f(\bm{f}_i)) + \log(b_p - d_p(\bm{p}_i, T)) & \text{if } d_p < \delta_T \\
-r_{1} & \text{otherwise}.
\end{cases}
\end{equation}
The contact-to-grasp phase 
finishes once all finger contact directions have converged within $\theta$.

In order to ensure that the grasp/contact phase switches, an avoiding penalty of $-r_{2} (< -r_{1})$ is applied if the phase does not switch from an approach phase within a certain number of iterations. 
Also, an avoiding penalty is added if the switch condition is violated (e.g., loosing contact between the 
finger and the object during the contact-to-grasp phase).

\subsubsection{other reward constraints}

During the training, we ensure that the object does not change much in position by adding a penalty if the object moves too much (the object can be slightly oriented however). In addition, the passive-form closure assumes that the grasp-object's (e.g., handle's) gravity is compensated by the environment. This means that the object is usually connected closely to some sort of environment (e.g., door). Therefore, we train the passive-form closure in a condition where an environmental plane is close to the grasp-object, and the fingers must go between the plane and the grasp-object.

In addition, the active/passive-force closure must evaluate whether enough contact force is being produced to resist against an external force. After finishing the contact-to-grasp phase, we evaluate the resistance against the external force that the grasp must resist against, and provide a $-r_{3}(>-r_{1})$ penalty on failure. The external force to resist against depends on the task sequence. For the experiment, we assume a pick which external force is gravity.

\subsubsection{domain randomization}

Domain randomization is done in simulation, and consists two stages. At every episode in the first stage, we simulate a "recognized contact-web" by adding random noise less than $5mm$ translation and $1.5^\circ$ rotation to the actual location of the contact-web in the simulator. Then, on a spherical surface which center matches the center of the "recognized contact-web," we randomly place the robot hand. This placing on the spherical surface is equivalent to randomizing the approach direction and can be represented with the azimuth and zenith angles. The state is represented respective to the "recognized contact-web" while the rewards are calculated using the contact-web's actual location in the simulator.

Once the agent no longer gets penalties, we do a second-stage randomization.
At every episode in the second stage, we further randomize the object shapes using superquadrics parameters~\cite{barr1981superquadrics} so that the policy can become robust against object shape uncertainty. The object scale and the shape of the object's upper surface are randomized within the range of the YCB object dataset~\cite{calli2015ycb}.


\section{Skill application in an LfO system}
\label{lfo}

\begin{figure*}[t]
      \centering
      \includegraphics[width=\linewidth]{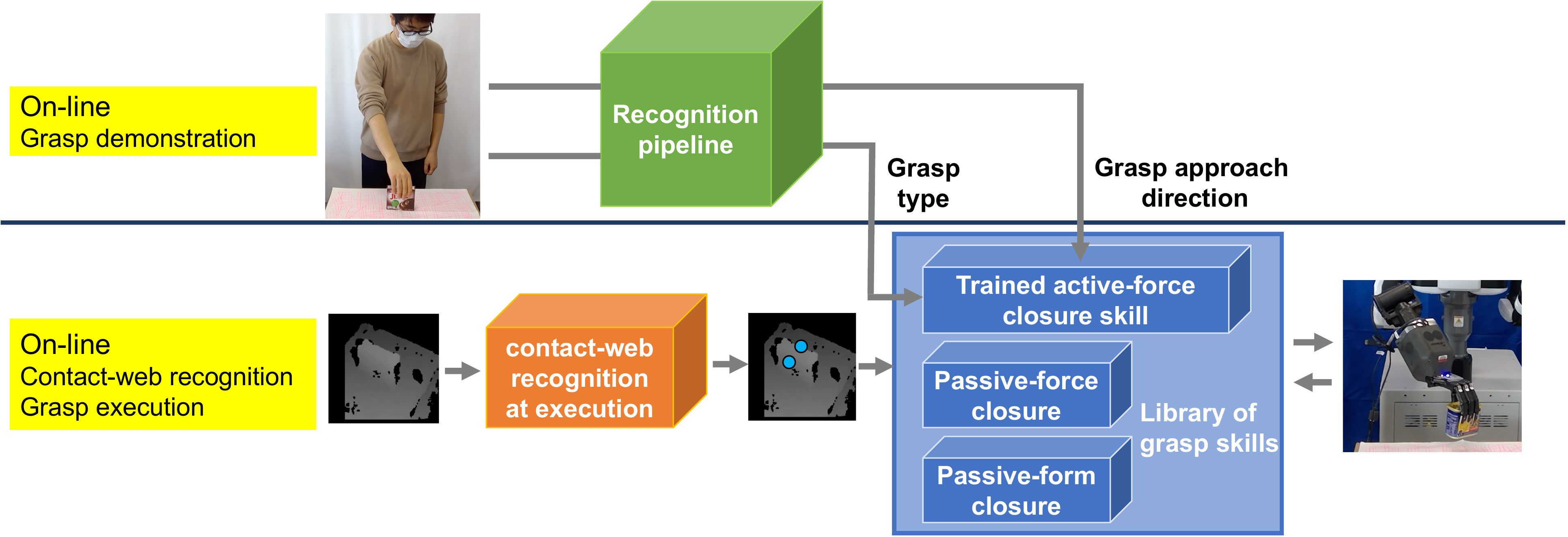}
	 \caption{The proposed LfO system overview during a grasping demonstration. Note that a human can demonstrate different tasks before and after the grasp, but the figure shows only the grasping part for the scope of the paper.}
      \label{fig:lfo}
  \vspace{-3mm}
   \end{figure*}

In this section, 
we explain how to achieve task-grasping by using the trained skills with the leveraging of the human demonstration.

The LfO system has two phases (\figref{fig:lfo}).
In the online grasp demonstration phase, the grasping skill appropriate for the task situation is recognized from the human-demonstration~\cite{wake2021verbal,wake2020grasp}.
The approach direction is also recognized~\cite{ikeuchi2018describing, wake2021learning} and is passed to the trained policy as a state\footnote{We expect that the object locations in the grasp execution phase to be roughly similar as in the grasp demonstration phase (we are mainly targeting a task-grasping in a human environment where we can assume the human stores/puts-back an object mostly in similar locations: a shelf, drawer, kitchen table etc.), therefore this approach direction is a reusable information for execution.}.

In the online grasp execution phase, 
the trained skill is applied to an arbitrary scenery by recognizing the contact-web locations within the depth image, and then the initial states are transformed respect to the recognized contact-web. 
From the initial state, the trained skill policy decides its first action, and then returns a new action from an updated state every 100 ms.


\section{Experiments}
\label{experiment}

\subsection{Training settings}
We trained the skills of three representative grasps (active-force closure, passive-force closure, and passive-form closure) using a robotic hand with necessary complexity. 
That is, simple-structured enough to experiment our concept on functionality mimicking grasps, but complex-enough to show that the training converges with a certain degrees-of-high-freedom.

We used the Shadow Dexterous Hand Lite which has four fingers, but we bind the index-middle-ring finger to bend together to produce simply-structured functional movement (two opposing virtual fingers), but still has $3$ degrees-of-freedom for the bind fingers, and $4$ degrees-of-freedom for the thumb.

We train with $N=2$ contact points for the active-force closure, with $N=3$ including the virtual fingertips and the palm for the passive-force closure, with $N=1$ (the virtual fingertip of the bind fingers) for the passive-form closure.

We used the proximal policy optimization (PPO)~\cite{schulman2017proximal}
to learn the policy
and used the Microsoft Bonsai Platform\footnote{https://preview.bons.ai/} to run multiple simulators in parallel for the training.
Each finger had a binary tactile feedback available in the states.

\subsection{Training results of the three grasping skills}

  \begin{figure}[t]
      \centering
      \includegraphics[width=\columnwidth]{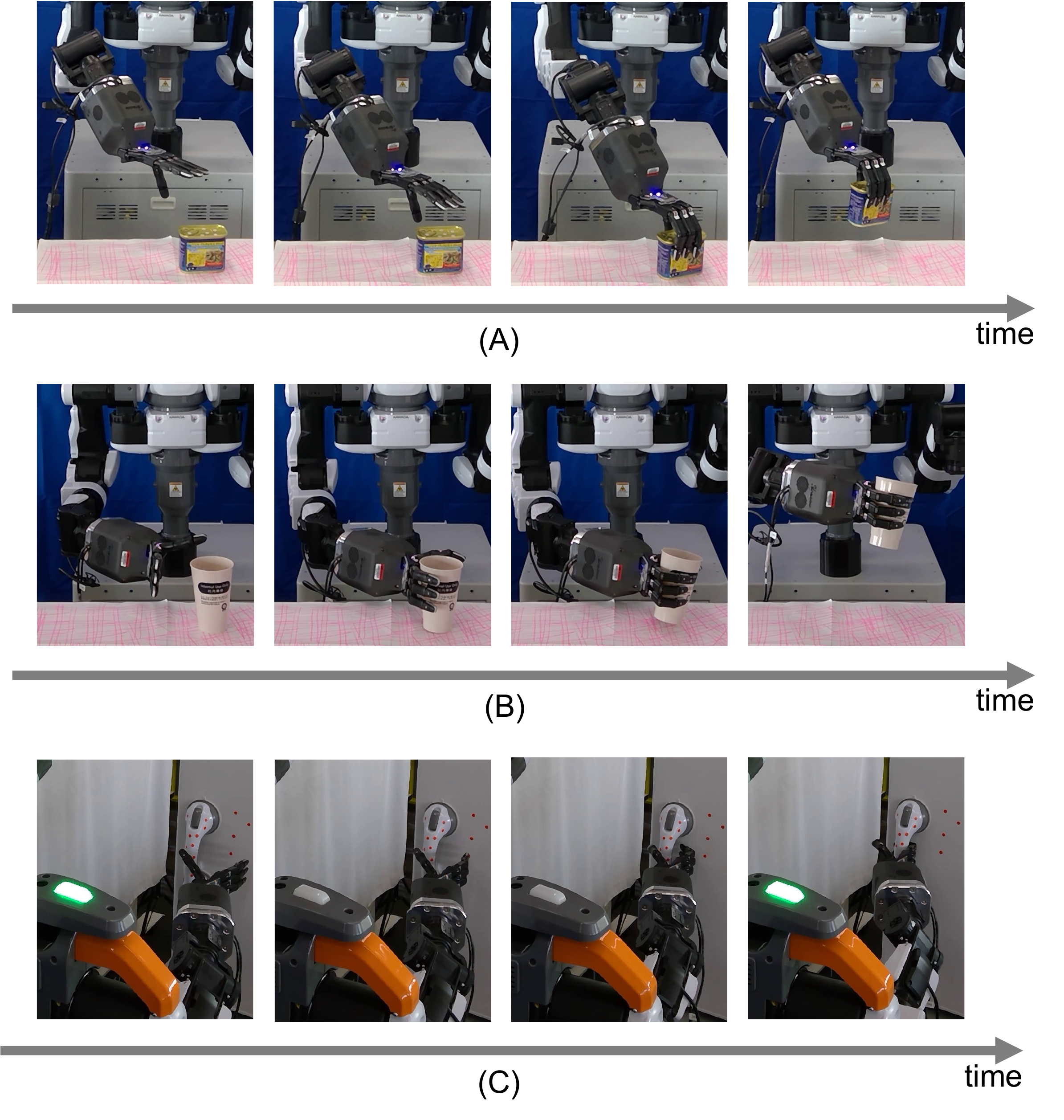}
	 \caption{Successful execution of the trained three grasping skills on the Shadow Dexterous Hand Lite. (A) grasp a box using active-force closure grasp. (B) grasp a cup using passive-force closure grasp. (C) hook on a door handle using passive-form closure grasp.}
      \label{fig:functional_grasping}
  \vspace{-3mm}
   \end{figure}

\figref{fig:functional_grasping} shows a successful execution of the trained three grasping skills on the real Shadow Hand Lite 
using the LfO pipeline described in \secref{lfo}.
The contact-web was recognized using a trained convolutional neural network (CNN). The CNN was trained using simulated depth images with ground truth labels using camera parameters from the real camera.
The recognition results had pose errors within the range of noise used in the policy training, and the robot did not have the scales and shapes of the object used in the human demonstration. 
The results indicate that, using the training design and domain randomization in simulation, a robust task-grasping skill can be executed with the real robot under recognition/object-model uncertainty.

\subsection{Robustness evaluation of the trained grasping skill}
\label{experimentb}

  \begin{figure}[t]
      \centering
      \includegraphics[width=\columnwidth]{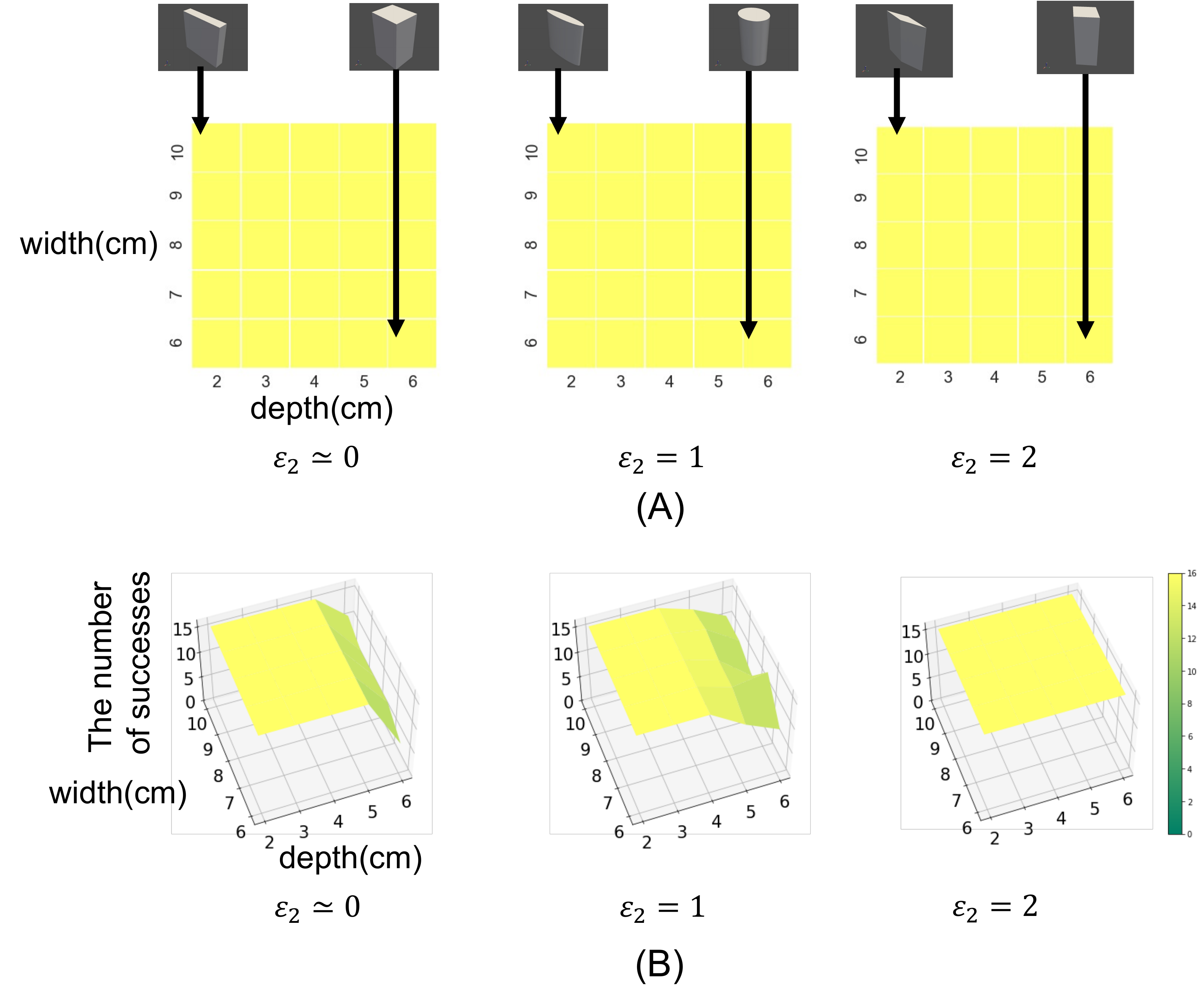}
	 \caption{Successful (yellow) and failed (green) grasps in each object shape for the active-force closure skill. $\epsilon_2$ represents the curvature of the top surface of the object. (A) Tested without object pose uncertainty. (B) Tested with 16 sampled pose uncertainty patterns.}
      \label{fig:robustness}
  \vspace{-3mm}
   \end{figure}

We evaluate how well the policy was trained in terms of robustness against object shape variance, which is critical if we would like to handle demonstrations using objects which the robot has no prior knowledge about (knowledge as in the precise model of the object). We focus on the policy on the active-force closure grasp, which is considered to be most difficult of the three 
representative
grasps.

The experiment was carried out in the simulator to investigate the robustness in detail. 
The approach direction was fixed to $(0^\circ, 0^\circ)$. 
For the scale and shape parameter, we evaluate whether the trained policy is capable of handling the 
range that we targeted for the skill (same as the range in the domain randomization in which we trained the skill).
Depth was changed by $1cm$ within $[2cm, 6cm]$ and width was changed by $1cm$ within $[6cm, 10cm]$, and $\epsilon_2$ was set to $0$ or $1$ or $2$.
Whether the grasp was successful or not was evaluated by lifting the object after finishing the grasp.
\figref{fig:robustness}~(A) shows the 
successful scales and shapes in yellow and failed in green.
\figref{fig:robustness}~(B) shows the robustness when recognition errors of the contact-webs are added on top, in which each object shape was grasped with 16 error patterns sampled from the range $5mm$ and $1.5^\circ$ noise range. 

The results indicate that
the policy was well-trained to grasp against 
a large range of objects (even under noisy conditions) despite being only a single trained policy. 
When recognition errors are added, the policy fails the thick and high-curvature shaped objects. The errors cause the fingers to easily collide with thick objects, and slipping-at-contact with high-curvature shapes.
This indicates the areas where grasping skills may require to train a sub-category based on the shape created by the fingertips (an encompassing shape instead of a parallel shape) to overcome "weak" geometries. However, such a strategy is only applicable in the case where the robot hand has more than two opposing fingers.

\subsection{Reach-coordination evaluation of the trained skill}

We show how well our active-force closure grasp was trained against approach variance. The experiment was done in a controlled simulation environment, and we chose three successful objects from Section~\ref{experimentb} but with different degrees of difficulty.
The first object (A) was the average size/shape object of the successful grasp (depth=4, width=10, $\epsilon_2 \simeq0$). The second object (B) was selected as a challenging-scale object with a small width (depth=4, width=6, $\epsilon_2 \simeq0$). The third object (C) was selected as a challenging-shape object with the upper surface shaped in a diamond (depth=4, width=10, $\epsilon_2$=2),
The approach
zenith was changed by $10^\circ$ within $[0^\circ, 60^\circ]$ and azimuth was change by $15^\circ$ within $[-45^\circ, 45^\circ]$. The recognition errors were not added.

\begin{figure}[t]
      \centering
      \includegraphics[width=\columnwidth]{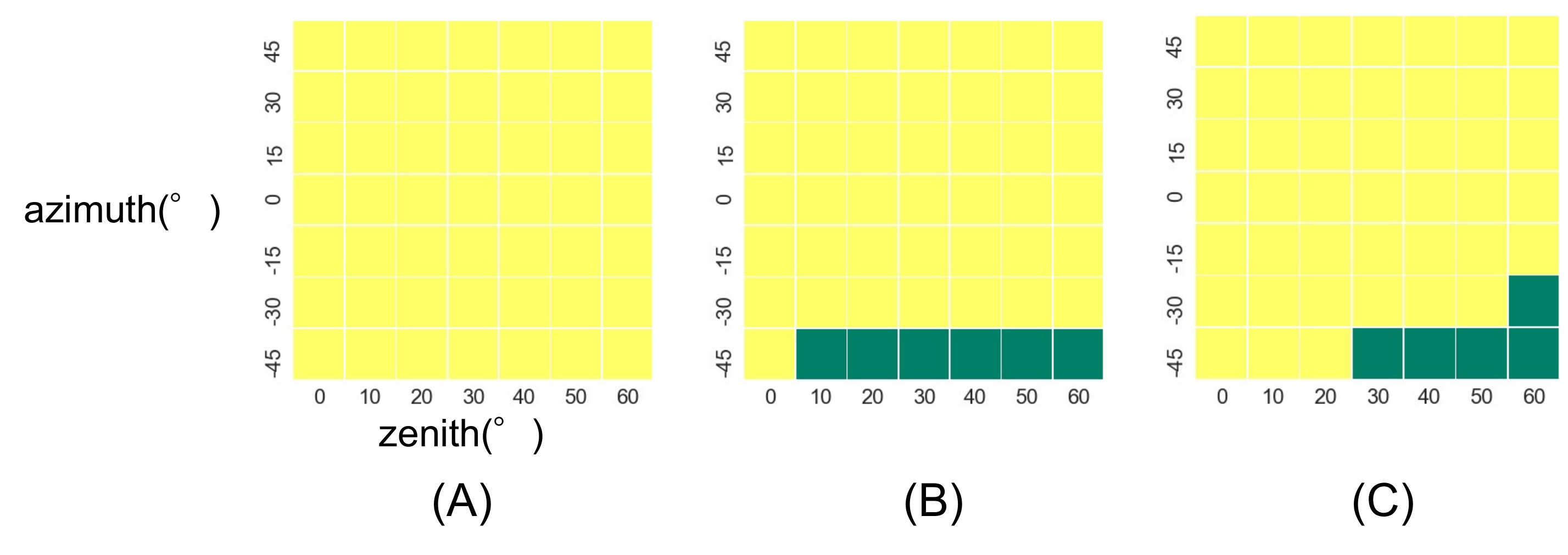}
	 \caption{Successful (yellow) or failed (green) grasps  with each approach direction in each tested object. (A) is the object which has (depth, width, $\epsilon_2$) $=$ ($4, 10, \simeq0$), (B) is the object which has ($4, 6, \simeq0$) and (C) is the object which has ($4, 10, 2$).}
      \label{fig:robustapproach}
  \vspace{-3mm}
   \end{figure}

\figref{fig:robustapproach} shows the
successful approach direction in yellow and failed in green. 
The object (A) was successfully grasped with all tested approach directions. The object (B) and (C) were successfully grasped with almost all tested approach, but were failed
when the azimuth was largely negative.
The asymmetric result on the azimuth is due to the slight asymmetrical structure of the robot hand used in the experiment, where the root of the thumb is located towards the robot's index-finger; the hand is specifically designed for the "right arm."
The hand structure makes the approaching of the thumb to the contact positions difficult for the largely negative azimuth, yet, this reaching direction is rare to happen during a task if executing with the "right arm," which the hand was designed for.
The success against approach directions is sensitive to the design of the hand.

\section{Conclusions}
\label{conclusions}
We proposed a method for training robust task-grasping skills using contact-web based rewards and domain randomization of approach directions. The skills can functionally mimic human-demonstrated grasping, and can coordinate with the reaching body to solve the LfO-based task-grasping problem.
Our results show that our trained policies are able to execute three representative grasps (active-force closure, passive-force closure, passive-form closure) on the real robot. We confirmed that the trained policy is able to grasp an object without the precise model of the object, under recognition error, and is able to grasp from various directions.

\ifconfletter
\bibliographystyle{IEEEtran}
\else
\bibliographystyle{unsrt}
\fi
\bibliography{bib}

%








\end{document}